\RequirePackage{etex} %% **************** add this for ARXIV submission
\documentclass[runningheads,twocolumn,a4paper,10pt]{llncs}

\usepackage{cite}

\usepackage[a4paper, total={7in, 10in}]{geometry}
\usepackage{booktabs}
\usepackage{graphicx}
\usepackage{amsmath}
\usepackage[hidelinks]{hyperref}
\usepackage{balance} %balance columns
\usepackage{lastpage}
\usepackage{authblk}
 \usepackage{xurl} %for long URLs
\usepackage{float}
\usepackage{bbding} %for star symbol

\newcommand\nnfootnote[1]{%
  \begin{NoHyper}
  \renewcommand\thefootnote{}\footnote{#1}%
  \addtocounter{footnote}{-1}%
  \end{NoHyper}
}

\begin{document} 

\title{Assessing Trustworthiness of Autonomous Systems}
\titlerunning{Assessing Trustworthiness of AS}
\authorrunning{Chance, Abeywickrama et al.}
\author{Gregory Chance\inst{1,2}  \and Dhaminda B. Abeywickrama\inst{1}  \and Beckett LeClair\inst{2}  \and Owen Kerr\inst{2}  \and Kerstin Eder\inst{1}}
%\author{Gregory Chance\inst{1,2}  \and Dhaminda B. Abeywickrama\inst{1}  \and Beckett LeClair\inst{2}  \and Owen Kerr\inst{2}}

\institute{Trustworthy Systems Lab, University of Bristol, Bristol, UK \and Digital Systems Assurance, Frazer Nash Consultancy, Bristol, UK}
\tocauthor{Authors' Instructions}

\maketitle
\nnfootnote{
%\textit{Statements about authorship contribution.}
%% Christopher Harper (e-mail: chris.harper@brl.ac.uk),
%% Saquib Alam (e-mail:saquib764@gmail.com), 
%% S\'everin Lemaignan (e-mail: severin.lemaignan@brl.ac.uk)
%% and
%% Tony Pipe (e-mail: tony.pipe@brl.ac.uk), 
%% are with the University of the West of England, Frenchay, Coldharbour Ln, Bristol, BS34 8QZ, United Kingdom.
%%
Greg Chance is the corresponding author (e-mail: g.chance@fnc.co.uk or greg.chance@bristol.ac.uk), 
%% Abanoub Ghobrial (e-mail: abanoub.ghobrial@bristol.ac.uk), 
%and 
%Kerstin Eder (e-mail: kerstin.eder@bristol.ac.uk) 
%are with the Trustworthy Systems Lab, Department of Computer Science, University of Bristol, Merchant Ventures Building, Woodland Road, Bristol, BS8 1UQ, United Kingdom.
}

\makeatletter
\renewcommand\subsubsection{\@startsection{subsubsection}{3}{\z@}%
                       {-18\p@ \@plus -4\p@ \@minus -4\p@}%
                       {4\p@ \@plus 2\p@ \@minus 2\p@}%
                       {\normalfont\normalsize\bfseries\boldmath
                        \rightskip=\z@ \@plus 8em\pretolerance=10000 }}
\makeatother

%\textbf{\textit{Abstract}--\input{\pathToSourceFiles/abstract.tex}}
\textbf{\textit{Abstract}--As Autonomous Systems (AS) become more ubiquitous in society, more responsible for our safety and our interaction with them more frequent, it is essential that they are trustworthy. Assessing the trustworthiness of AS is a mandatory challenge for the verification and development community. This will require appropriate standards and suitable metrics that may serve to objectively and comparatively judge trustworthiness of AS across the broad range of current and future applications. The meta-expression `trustworthiness' is examined in the context of AS capturing the relevant qualities that comprise this term in the literature. Recent developments in standards and frameworks that support assurance of autonomous systems are reviewed. A list of key challenges are identified for the community and we present an outline of a process that can be used as a trustworthiness assessment framework for AS. 
%
% We conclude that this is a really difficult problem and no one should try it.
}

% ***************************************************
%  Main Body
% ***************************************************
%\input{\pathToSourceFiles/body.tex}
% \addbibresource{../source/TASverif.bib}

% ****************************************
% *************************** Introduction
% ****************************************

\section{Introduction}\label{sec:intro}

Autonomous systems are systems that involve software applications, machines and people, which are capable of taking actions with no or little human supervision~\cite{Murukannaiah2020}.
Autonomous systems (AS) are pervasive in current society and set to become even more so with current technological growth trends and adoption rates. Systems with embedded artificial intelligence (AI) and machine learning (ML) algorithms can be found in numerous applications from chatbots like GPT\-3~\cite{floridi2020gpt}, mobile phones~\cite{mediumaiphones}, insurance pricing models~\cite{kuo2020towards}, vacuum cleaners~\cite{tf_vacuum} and self-driving vehicles to medical diagnostics~\cite{kononenko2001machine}, detecting structural damage to buildings~\cite{avci2021review} and predicting the shape of protein molecules~\cite{alpha_fold} to name a few.
For successful adoption and reliance upon AS, there needs to be demonstrable assurance of their trustworthy operation which becomes increasingly difficult for complex systems with more ambitious \emph{automation scope} and being used for applications with greater criticality. 
%
% There is also growing use of machine learning in a range of safety-critical systems (SCSs), for example in the aerospace and automotive industry where low reliability of these systems could result in catastrophic failure and potentially loss of life or damage to property and the environment. These safety-critical autonomous systems present a complex but essential challenge to the safety assurance and verification community. 

Verification and validation (V\&V) is the process to gain confidence in the correctness of a system relative to its requirements. Prior to, and separate from verification, a specification must clearly define the trustworthy operational behaviour of the system and many challenges are associated with this task for autonomous systems~\cite{Abeywickrama2022}. 
%
% Conventional V\&V is principally concerned with assessing the system against a set of requirements, providing guarantees of functionality and assurance of safety.
%
If autonomous systems are to be fully trusted into society, there must be acknowledgment of, and evidence to show, compliance with a broad range of \emph{trustworthiness qualities}. 
A trustworthiness quality is defined here as a non-functional system property that promotes reliance or enables adoption in that system with respect to the views of the \emph{trust stakeholders}. A survey of the literature on trustworthy AS is presented and has identified many qualities that have been grouped into an ontology including properties such as reliability, robustness, security, accountability and ethics. 
Trust stakeholders have a vested interest in the reliable operation of the system or otherwise seek assurance of specific system properties from a variety of perspectives, such as: end users or operators, regulators, developers or system manufacturers, unintended users and, to some extent, the environment. 

A major challenge in verifying this broad category of trustworthiness qualities, is the availability of standards and regulations against which they can be evaluated. Whereas verification methods of assessing (for example) functional correctness are relatively mature, there also exists the challenge of developing robust assessment methodologies and metrics for non-functional and nuanced trustworthiness qualities, e.g. fairness and beneficence. Additionally, there may be a new class of AS possessing evolving functionality, where functionality automatically changes over time - presenting additional challenges to verification and standards. 
Some commentators argue that for complex systems with high automation scope, the level of proof required for verification at design time is intractable and that some form of runtime verification is required~\cite{althoff2014online,CyRes20}. In addition, systems with operational adaptation or evolving functionality should also include some level of runtime monitoring of appropriate metrics to ensure the system does not move to an untrustworthy state or condition. 
%
% [maybe cut] For example, there are well defined rules for driving conduct and expected behaviour [UKHC] which could be verified using simulation based verification [harper22] but standards for social interaction, aesthetics or ethical behaviour are either non-existent or just emerging [EU directive for AI ethics]. 

This paper contributes in the following areas: reviewing what trustworthiness means in the field of AS, robotics, Human Robot Interaction (HRI) and Cyber-Physical Systems (CPS) in both the academic literature and any standards or relevant frameworks; how the application, criticality and automation scope influence trust assessment; presenting an assessment framework that can be used as a basis by which AS can be verified. 

Trustworthiness \emph{of} the system must also be reciprocated with user trust \emph{in} the system for successful adoption and subsequent reliance~\cite{Lee2004}. User trust may consider other contributing factors such as the broader epistemological or sociotechnical aspects of trust, but these will not be considered here, nor the process of gaining, maintaining or preventing the erosion of the \emph{trust agreement} between user and system, of which there are many excellent discussions, e.g. see~\cite{Kohn2021,Lee2004,kok2020trust,Chiou2021,Floridi2018}. 
%
%We focus on technical aspects of objective trustworthiness relating to the system and outline the challenges associated with practical application of a presented assessment process. 

%This document is structured as follows: ...

% ****************************************
% ***************************** V&V for AS 
% ****************************************

\subsection{Verification and Validation for AS} \label{sec:intro-vav}

Attaining complete assurance of any complex system is challenging and, in some cases where the input parameter space is large, it may be intractable. 
However, there are multiple techniques and approaches that the verification engineer can use to seek assurance against certain system properties, such as formal modelling, testing, synthesis and runtime verification~\cite{kress2021formalizing}. 
These techniques and the V\-model of verification have been used to successfully provide assurance of software functionality for several decades~\cite{Fewster1999}. 
But as the functionality, ambition and level of automation of deployed systems grows, the assessment methodology must also adapt and account for these new responsibilities. This may include aspects such as security, privacy and other ethical concerns, highlighted by, for example, the need for better online security and protection against web search surveillance~\cite{trackmenot2009}. 
There are emerging developments in the areas of ethical AI assessment and assurance, for example Porter et al. propose a principles-based ethical argument (PBEA) framework for reasoning about the overall ethical acceptability of an AS~\cite{Porter2022}.

% A widely held tenet is that there can never be a suitable amount of verification that gives complete assurance for complex, safety-critical autonomous systems, although limits on reliability rates have been proposed~\cite{Butler1993}. 

Corroborative V\&V~\cite{webster2020corroborative} attempts to improve confidence through combining mutually consistent evidence from multiple and diverse assessment methods, e.g. formal, testing~\cite{schwamm2022}. 
But even this may not be enough for the diverse operational domains of some AS and thinking should move beyond design time verification, to a more continuous operational evaluation such as \emph{runtime verification}. Runtime verification brings other currently unresolved issues, such as suitable oracle design~\cite{Leucker2009}, but some authors propose valid ideas to this using edge computing as a cloud-based verification authority~\cite{CyRes20,eder2021cyres}. 
%
% The use of \emph{serious games} can be another interesting opportunity for building trustworthiness in complex autonomous systems and has been used in the context of mission planning for NASA~\cite{Allen2018} and in a self-driving vehicle controller leveraging the power of crowdsourcing for test generation [ref test gen game, need to make github page, add code and youtube videos]. 

Further to these issues are the lack of standards against which some trustworthy qualities should be appraised and the methods by which they should be evaluated. For example, there are standards for correct road driving conduct~\cite{highwayCode} but ethical standards by which those driving decisions should be made do not exist or are just emerging~\cite{Bonnefon2016} . 
An interesting dilemma is if maximising the trust in qualities results in conflict, and how these conflicts can be ethically resolved. 
Although headway is being made into developing standards for non-functional properties, such as guidelines for ethical AI~\cite{Floridi2018}, checklists for HRI best practice ~\cite{kraus2022trustworthy} and transparency~\cite{winfield2021ieee}, there are still areas that need attention such as standards for adaptability, cooperation and fairness~\cite{Abeywickrama2022}. 
Where standards are lacking or immature, this will require engagement with \emph{trust stakeholders} - expert steering groups that can define and prioritise the necessary trustworthiness qualities for each subject domain or application. 

Additionally, there is more that can be done at the design stage to improve \emph{verifiability}~\cite{kusters2010}. Evidence for functional correctness is essential, but this must be supported with decision explanation~\cite{koopman2018toward} whilst maintaining intellectual property rights around, for example, sensitive software algorithms and trade secrets. 

In addition to assessing AS trustworthiness, there must also be consideration to gain, calibrate and maintain user trust in the system~\cite{kok2020trust, Chiou2021}, as miscalibration of trust between system and user can have serious consequences~\cite{kok2020trust}, although this is not the focus of this work.

%\subsection{Document Structure}
% In the following, related work is reviewed in Section

% ****************************************
% ************** Trustworthiness Qualities
% ****************************************

\section{AS Trustworthiness Qualities}\label{sec:astq}

As computing and automation has developed, systems are now both more capable and users more reliant on them for productivity in several sectors. This extension of capability has resulted in a broadening of the terms which encompass trustworthiness, as, for example, the important trustworthiness qualities of a calculator may be less numerous than those of a medical decision support system. Advancement in automation then, has led us to question and challenge these new capabilities, or, as some commentary has noted: with more automation comes more responsibility~\cite{Yazdanpanah2021}. 

Trust can be expressed in a number of ways and directions; trust the user has in the system, the objective trustworthiness of the system and the context in which the interaction between the two takes place~\cite{Hancock2021}. 
Trustworthiness can also been described as a the probability that a system holds some established property or quality, and that greater trustworthiness begets greater likelihood that the system may exhibit that quality. 
In this research we consider the trustworthiness of the system and the specific qualities that must be demonstrated, but we acknowledge the importance of the other mechanisms where human-system trust can be gained or lost in which there has been much contribution from the HRI, psychology and human factors community~\cite{Floridi2019,Lee2004,kok2020trust,Chiou2021,Kohn2021,kraus2022trustworthy}. 
Trustworthiness of autonomous systems in the context of this work therefore results from objective assessment of the system with respect to a set of appropriate standards. 
There has been much academic deliberation on the specific qualities that comprise trustworthiness of AS, specifically for AI~\cite{Thiebes2021,Wing2021} and HRI~\cite{kraus2022trustworthy,atkinson2012trust}. 
Devitt argues that reliability and accuracy are the two central pillars of trustworthiness of AS and that all other properties stem from these, for example, stating that adaptability and redundancy are higher-order properties of reliability~\cite{devitt2018trustworthiness}. 

The Trustworthy Software Foundation~\cite{ts_foundation} state 5 facets of trustworthy software: Safety: The ability of the software to operate without causing harm to anything or anyone; Reliability: The ability of the software to operate correctly; Availability: The ability of the software to operate when required; Resilience: The ability of the software to recover from errors quickly and completely; Security: The ability of the software to remain protected against the hazards posed by malware, hackers or accidental misuse.

% ****************************************
% ************************ Ontology of AS
% ****************************************

\subsection{Ontology of AS Trustworthiness Qualities}\label{sec:tasq-ont}

A trust ontology can be a useful definition to identify an independent set of important quality characteristics, where one category is not necessary influenced or related to its neighbours. These categories can be used to support clarity of communication and understanding of issues pertaining to, and of judgement in the assessment of, trustworthiness of autonomous systems. 

Lee \& Moray propose the categories for trust in automation: performance (consistent and stable behaviour), process (qualities or characteristics that govern behaviour), purpose (underlying motive or intent) and foundation (fundamental assumptions of natural and social order). These categories broadly capture the full gamut of trustworthy qualities but may be too broad and abstract for practical assessment purposes. 

Avizienis et al. proposes that a set of general concepts are required for dependable and secure computing, which may cover a wide range of applications and system failures comprising: availability (readiness for correct service), reliability (continuity of correct service), safety (absence of catastrophic consequences on the user and the environment), integrity (absence of improper system alterations) and maintainability (ability to undergo modifications and repairs)~\cite{avizienis2004basic}. The focus here is on functionality and usability, but these categories may be too specific to computing and neglect the verifiability and ethical considerations surrounding AI.

Thiebes et al. argue for five foundational principles of trustworthy AS: beneficence (doing good), non-maleficence (doing no harm), autonomy (preserving human decision making), justice (being fair and reasonable), and explicability (being easily understood)~\cite{Thiebes2021}. These are based on and related to numerous other discussions on ethically-principled foundations of trustworthiness; there is evidence of strong international collaboration and motivation in this area~\cite{Floridi2018,jobin2019global}. Whilst these categories are very important and capture ethical and regulatory considerations, they fail to capture the aspects of functionality and dependability of other voices in the community.

Cho et al. propose a STRAM ontology for measuring the trustworthiness of computer systems, based around four sub-metrics of: security (availability, confidentiality, integrity), trust (predictability, safety, reliability), resilience (adaptability, fault-tolerance, recoverability) and agility (efficiency, usability, timeliness), although again functional aspects are missing with the main focus being on security.

The National Institute of Standards and Technology (NIST) produced a set of four guiding principles for explainable AI~\cite{nist4}. While useful in its own right, this document also outlines their ontology for trust, which comprises explainability alongside interoperability, accuracy, privacy, reliability, robustness, security, safety, bias mitigation, transparency, fairness and accountability. 

In addition to objective trustworthiness of AS, there should be consideration given to the governing principles of the people and companies that invent and develop these technologies, and the impact that these may have on their products and subsequently the people who use them. Keymolen outlines three key `pillars' of trust - competence, management commitment, and the provision of reasonable assurance~\cite{keymolen}, and argues that companies should nurture techno-moral competencies that go beyond mere legal compliance.  

%These pillars are somewhat less clearly defined than those given by other authors, though they are aimed more at the average corporate stakeholder who may not have as in-depth an understanding of AI technical aspects as the development teams they work with.

% ****************************************
% ********************* Existing Standards
% ****************************************

\begin{figure*}[t]
	\centering
\includegraphics[width=0.98\linewidth]{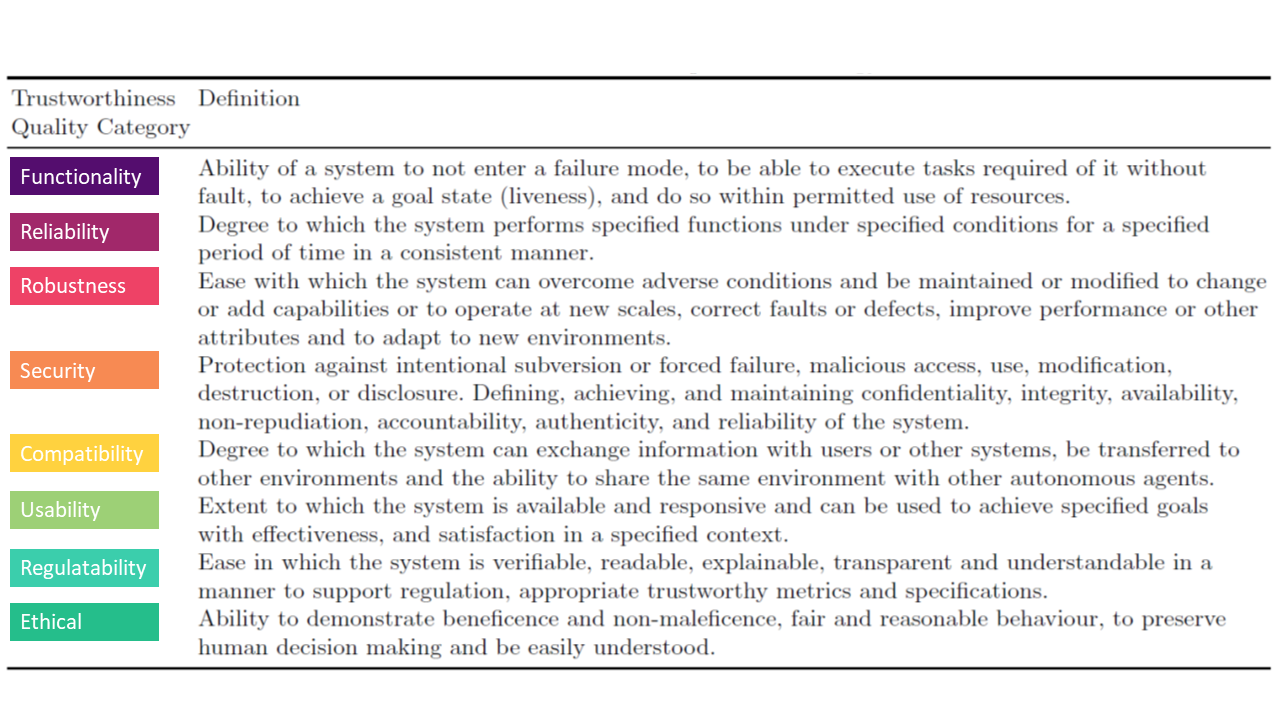}
	\caption{Trustworthiness categories for Autonomous Systems.}
	\label{fig:quals}
\end{figure*}

\subsection{Standards for Autonomous Systems and Trustworthiness Properties} \label{sec:standards}
% Contributed by: Dhaminda Abeywickrama
% 2 September 2022
% Autonomous systems are systems that involve software applications, machines and people, which are capable of taking actions with no or little human supervision~\cite{Murukannaiah2020}. %moved up to intro
%The functionality of an autonomous system (i.e. what it is meant to do, what it does, and what it could do) may evolve or change over time. 
One of the main issues of adopting current standards and regulations with autonomous systems is the lack of consideration to the notions of \textit{uncertainty} and \textit{autonomy} \cite{Fisher2021}. 
Most conventional processes for defining system requirements assume that these are fixed and can be defined in a complete and precise manner before the system goes into operation \cite{Abeywickrama2022}. 
Also, in existing standards and regulations, the notion of autonomy is not their most characterizing feature where they are neither driven nor strongly influenced by it \cite{Fisher2021}. 
Most existing standards are either implicitly or explicitly based on the V\&V model, which moves from requirements through design onto implementation and testing before deployment~\cite{Jia2021}. 
However, this model is unlikely to be suitable for systems with the ability to adapt their functionality in operation; e.g.\ through interaction with other agents and the environment (e.g. as is the case with swarms); or through experience-driven adaptation as is the case with machine learning \cite{Abeywickrama2022}. 
%
%\textcolor{red}{we have not yet mentioned systems with evolving functionality, this can be put in the introduction} - added
%
Autonomous systems with evolving functionality follow a different, much more iterative life-cycle. Thus, there is a need for new standards and assurance processes that extend beyond design time and allow continuous certification at runtime~\cite{Rushby2008}. In this context, there have been standards and guidance introduced by several industry committees and research groups. Below we give an overview of several key efforts with any trustworthiness properties or ontologies supported by them.

In 2016, the British Standards Institution introduced the \textit{BS 8611} standard that provides a guide to the ethical design and application of robots and robotic systems \cite{BS8611}. 

IEEE, through its  Global Initiative on Ethics of Autonomous and Intelligent Systems, developed a series of standards (IEEE P70XX) to address autonomy, ethical issues, transparency, data privacy and trustworthiness. The \textit{IEEE P7001} standard describes measurable, testable levels of transparency for autonomous systems so that they can be objectively assessed and levels of compliance can be determined~\cite{IEEE-P7001}. This standard outlines five stakeholder groups, and for each group it explains the structure of the normative definitions of levels of transparency. IEEE P7001 can be applied to assess the transparency of an existing system using a process of System Transparency Assessment, or to specify transparency requirements for a system prior to its implementation using a System Transparency Specification.

Additionally the \textit{IEEE P7007} standard aims to assist in ethically-driven methodologies for the design of robots and automation systems \cite{IEEE-P7007}. For this, it provides a set of ontologies with different abstraction levels of concepts, definitions, axioms and use cases. 

IEEE P7001 discusses the transparency concern, which is a property representing an explanation topic (e.g. fairness, safety, legality, reliability, accountability, responsibility, predictability, comprehensibility, justifiability, viability, coordination) describing the reason for explanations of agent behaviours. Since the draft was proposed, this has gone on to be ratified in the Standard for Transparency of Autonomous Systems~\cite{FinalP7001}.

The \textit{IEEE P7010} standard is used to measure the impact of AI or autonomous and intelligent systems on humans~\cite{IEEE-P7010}. Recently, ISO/IEC TR 24028:2020 has also been introduced, aiming to provide a more general standard on how to define trust in the context of AI~\cite{24028}. ISO also publish standards which delve deeper into specific applications, for example 24029 which looks at robustness in neural networks~\cite{24029}.
A more general standard also exists for recording transparency in the algorithms used to drive ML and AI systems, courtesy of the Centre for Data Ethics and Innovation within the UK government~\cite{cdei}.

There are several standards and guidance pieces related to machine learning in aeronautics, automotive, railway and industrial domains, e.g. European Union Aviation Safety Agency (EASA) concept paper, UL 4600 standard \cite{Kaakai2022}. General guidance also exists in the form of the recently-published NIST AI Risk Management Framework~\cite{airmf}.

\textit{Assurance of Machine Learning for use in Autonomous Systems (AMLAS)} provides guidance on how to systematically integrate safety assurance into the development of machine learning components based on offline supervised learning \cite{Hawkins2021}. 
%AMLAS provides an explicit and structured safety case that the system is safe to operate in its intended context of use. 
AMLAS contains six stages, and the assurance activities are performed in parallel to the development of machine learning component. The process is iterative by design and feedback is used to update previous stages. In AMLAS, the safety requirements are always based on performance and robustness of the machine learning model. In work related to AMLAS \cite{Ashmore2021}, the authors identify several phases in a machine learning life cycle (data management, model learning, verification and deployment) with their associated data. From an assurance view point, they consider several key properties the models generated by learning should exhibit: performance, robustness, reusability and interpretability.

\begin{figure*}[]
	\centering
	\includegraphics[width=0.98\linewidth, trim=2cm 4.5cm 2.5cm 2.5cm, clip]{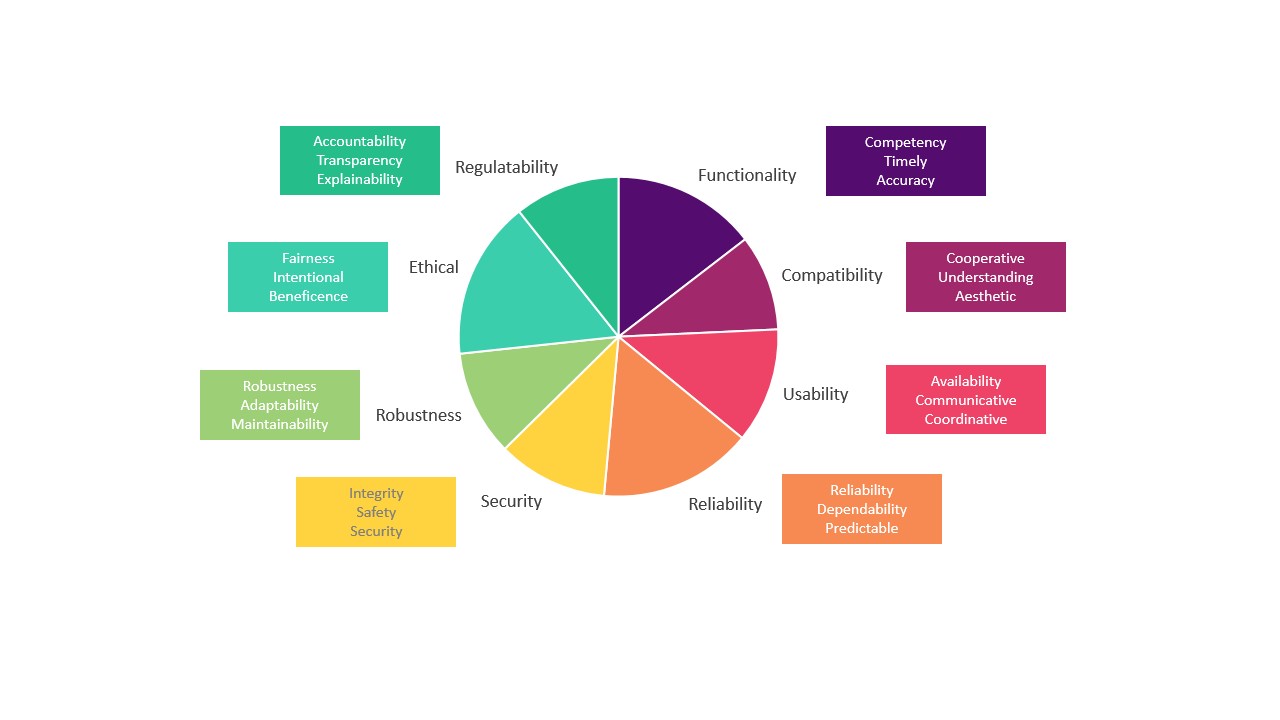}
	\caption{Analysis of trust quality terms in the literature placed into categories, breakout box shows most cited words from each category.}
	\label{fig:trust_spectrum}
\end{figure*}

The \textit{concept paper} by the EASA provides its first usable guidance for Level 1 (human assisted) safety-related machine learning applications \cite{EASA2021}. 
%It provides means of compliance for the certification of safety-critical systems which rely on machine learning-based algorithms for their operation. 
This guidance provides a roadmap to create a framework for AI trustworthiness (\cite{EASA2021}, pg. 8). The framework describes three techniques for analysing trustworthiness (safety, security and ethics-based), which are linked to the ethical guidelines developed by the EU Commission (accountability, robustness, safety, oversight, privacy and data governance, non-discrimination and fairness, transparency, and societal and environmental well-being). 

The \textit{DEpendable and Explainable Learning (DEEL) white paper} aims to identify challenges in the certification of systems using machine learning and to define a set of high-level properties for that purpose, such as auditability, data quality, explainability, maintainability, resilience, robustness, specifiability and verifiability (\cite{Mamalet2021}, pg. 22–23). 

The \textit{Aerospace Vehicle System Institute (AVSI) report} on machine learning summarises their findings on safety and certification aspects of emerging ML technologies that are applied to safety-critical aerospace systems \cite{AFE2020}. This report provides several recommendations with respect to robustness, safety assurance, runtime assurance and interpretability when using machine learning in safety-critical applications. 

The \textit{UL 4600 standard} guides a user through the development of safety cases for fully automated vehicles (i.e. vehicles with no driver or supervisor) \cite{UL4600}. It is more of a standard of care and not a procedure for certification of fully automated vehicles. The \textit{Laboratoire National de Métrologie et d'Essais (LNE) certification} \cite{LNE2021} is a quality assurance standard for machine learning processes. The aim is to provide guidance for an applicant when obtaining certifications for their design, development, evaluation and maintenance in operational conditions. 

Fenn et al.~\cite{fenn2023architecting} discuss how to apply safety and safety assurance, using common architectural patterns in safety critical aviation systems, to systems with AI/ML functionality. Although not a standard, this paper outlines work at KBR/NASA relating to the work progressing on the SAE G-34 AS6983 airworthiness standard for systems implementing AI. 

Hawkins et al.~\cite{Hawkins22} have presented a comprehensive guidance framework on the safety assurance of AS in complex environments (SACE) covering operational domain models, hazardous scenario identification, requirements definition and design assurance, failure mitigation and how an AS can operate outside of the usual domain boundary. This work offers an extension to Stage 8 in the SACE framework covering verification assurance. 

%\subsubsection{Ontologies within Existing Standards}\label{sec:tasq-ont-stds}

The international standard ISO/IEC/IEEE 29119 describes software and systems engineering and Part 4 covers software testing techniques and outlines 8 areas that testing should focus around: Functional Stability, Performance Efficiency, Compatibility
Usability, Reliability, Security, Maintainability, Portability~\cite{ISO29119}. 
This standard is primarily focused on software testing and so some of these categories, although useful, include jargon specific to computing systems which were not repeated in any other literature pertaining to AS more generally, and are therefore less user-friendly. However, Part 13 of ISO/IEC/IEEE 29119 sets out standards specifically for testing AI-based software systems, which extends these quality characteristics to include AI-specific qualities such as: flexibility (range of behaviours), adaptability (ease of modification or achieving flexibility), autonomy (unsupervised ability and level of control), evolution (behaviour adaptation over time), bias (e.g. due to discrimination, historic bias, or uneven sampling), transparency (access to data and algorithms and decision interpretability), and determinism (same output for given input) as well as consideration to ethical specifications and side-effects such as reward hacking. 

%FUTURE:ISO standard on "quality model for AI-based systems"

%DIN SPEC 92001 is a standard to help ensure quality in AI systems. Part 2 of the standard (92001-2) describes three pillars responsible for AI quality, namely: functionality and performance, robustness and comprehensibility - look into

% maybe drop
%The NIST Framework for Cyber-Physical Systems [ref] details a list of trustworthy `aspects' and `concerns' in addition to operational and business concerns for CPS and also includes some excellent case studies to show a complete end-to-end analysis, whilst Balduccini goes on to draw reasoning about the trustworthy properties set out in the framework in a UML/XML language [ref].  
%

% notes:

% Safety of autonomous systems (from Hawkins 2022): UL4000 [Underwriters Laboratories. Standard for evaluation of autonomous products, 2020] or SCSC-153B [Safety of Autonomous SystemsWorking Group. Safety assurance objectives for autonomous systems, 2022. URL: https://scsc.uk/scsc-153B]

\subsubsection{Trustworthiness Spectrum}\label{sec:spectrum}

A review of over 60 academic papers and standards around trustworthiness of AS, HRI and CPS was conducted as part of this work, see~\cite{tsl_git}. The aim of this review was to understand the expected properties of a trustworthy system that is automated or has integrated ML. The properties described by each paper were then collated into one of eight categories making up a spectrum of trustworthiness.

For each property described in the literature review semantic themes were investigated and collated, e.g. reliable, stable, predictable and reputable are semantically linked terms to describe reliability. These terms were then counted for frequency of occurrence and binned into the 8 themes in the spectrum.

The spectrum of qualities that captures the broad definition of trustworthiness of AS from the literature is shown in 
%Table~\ref{tab:quals}
Fig.~\ref{fig:quals}
. Reading from top to bottom, the trustworthy categories in the first column begin objectively and become increasingly subjective moving down the list. The spectrum of qualities comprises all of the trustworthy properties reviewed in the literature and the second column of 
%Table~\ref{tab:quals} 
Fig.~\ref{fig:trust_spectrum} 
gives a short description of the category in more detail. A full list of the quality terms reviewed and the semantic themes can be found at~\cite{tsl_git}.

\section{Key Considerations for Trustworthy AS} \label{sec:key}

The following section discusses other considerations for AS against which the trustworthy qualities should be examined. 

% ****************************************
% **************** Application Criticality
% ****************************************
\subsection{Application Criticality} \label{sec:appcrit}

What is not considered a great deal in the literature is application criticality - what application the AS is used for and if this should change the significance of specific trustworthiness qualities. Applications will need more emphasis on certain trustworthy qualities depending on where the system is most vulnerable to violating trust. 
%
%\textcolor{red}{Arianna, is there a focus in ethics on where the power is held? and whom can be prejudiced/discriminated against?} 
%
For example, a self-driving vehicle, or indeed any safety-critical system, must have emphasis on safety, possibly to the detriment of other qualities. 

DIN SPEC 92001-1 describes a \emph{quality meta model} and distinguishes between high risk systems that have safety, security, privacy and ethical relevance and those that do not (low risk), delineating applications into two risk classes~\cite{Englisch2019} which can be assessed using an appropriate risk assessment process from engineering (e.g. FMEA) or sociotechnical disciplines~\cite{macrae2021learning}. High risk applications must commit evidence of system trustworthiness based on these categories (or must be justified) whilst low risk systems are less strict. 

Whilst this DIN SPEC approach is commendable, it does not go far enough to filter trust qualities based on the application and identify the key qualities required for assessment. The risk assessment process can be used to identify those qualities which are most pertinent to the application which can be prioritised, included or discarded entirely. For example, there may be an application with strong safety requirements but little to none regarding privacy. 

Fisher et al.~\cite{Fisher2021} asks if there are some rules that are more important that others, that is, are all rules born equal? Context of application and sociocultural norms will influence the answer to this question. An example may be the ease with which breaking the speeding limit is observed in driving behaviour, but other driving conduct rules are broken less often, such as driving through a red traffic signal light.

\subsubsection{Potential harm (from failure)} \label{sec:app-harm}

Qualities that are principally connected to the functionality of the AS are required for trustworthiness. A small vacuum cleaner with poor coordination can do limited harm to users, but the same lacking quality in a larger robotic assistant, say, may fail to be accepted as trustworthy and also cause potential harm. Therefore, the trust quality must be elevated to a higher risk level. 
A risk assessment process should be able to identify such critical properties that can follow through the verification process.

% ****************************************
% *********************** Automation Level
% ****************************************
\subsection{Automation Level} \label{sec:autlev}

The level of automation is an important consideration which relates to what qualities the system should present, our reliance and vulnerability to the system and hence the criticality of the application. A good description of automation levels is given by SAE International~\cite{SAEJ3016}. 
Fisher et al. describe \emph{automation scope} which, alongside the level of automation, describes the sophistication of potential system actions and the ability to achieve complex task goals~\cite{Fisher2021}. 
Alongside scope, agency (independent acting or decision making) and whether to be reactive (responding to a situation or stimulus) or proactive (creating a situation) in action decision are also factors to consider within automation level.
%
%Greater scope, greater responsibility [ref]
%
% The example given is that two automated vacuum cleaners can both clean and avoid obstacles, but one schedules action to avoid human disruption. 
%
It could be seen that automation scope is an aspect of some non-functional AS qualities such as usability and compatibility, which includes co-existence and harmony; these are aspects of automation that would naturally extend the scope of the system.

\subsubsection{Decision Making Complexity} \label{sec:appcrit-dec}

Within automation level comes decision making complexity, the complexity of the process the AS must navigate in order to achieve the task goal. This could be considered in terms of the system and the constraints in the environment or action space to allow or prevent the system reaching a number of potential future states. An autonomous vacuum cleaner, for example, is physically constrained to a small 2-dimensional plane (area to be cleaned) and can successfully function with a small action space (stop, rotate, drive) where decisions are reactively made based on a few simple sensor inputs (avoid close object). 

Contrast this to, for example, any system that requires a perception stack to interpret dynamic physical scenes, such as a self-driving vehicle, which must identify and extrapolate objects and their future states (pedestrians), environmental conditions (road furniture, fog) and static or temporary rules that require interpreting (road signs, traffic cones) which will all contribute to a decision, which is also exacerbated by the potential harm that can result if a wrong decision is made. 

Decision making complexity needs to be considered when judging the risk level of the application, and the associated trust qualities should be elevated accordingly. Where applications have high risk associated with particular qualities, those qualities should be elevated to higher risk levels and be assessed accordingly (see Section~\ref{sec:AssFramVis}).

\begin{figure*}[t]
	\centering
	\includegraphics[width=0.98\linewidth]{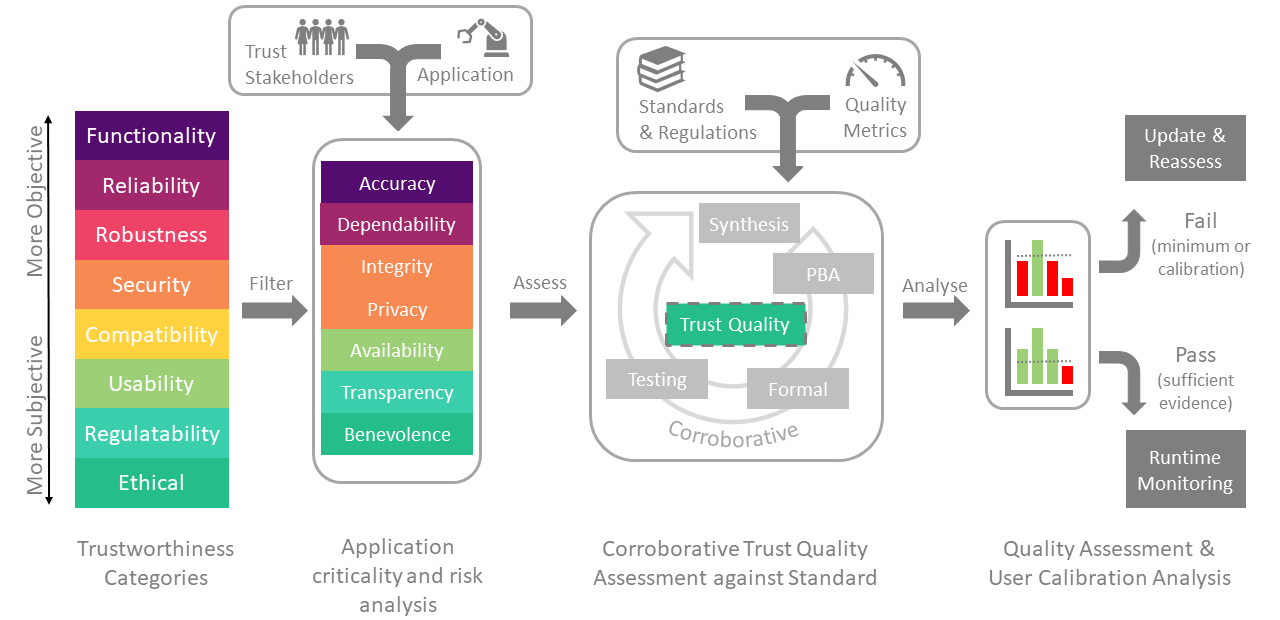}
	\caption{AS trustworthiness assessment process}
	\label{fig:tas_ver}
\end{figure*}

% ****************************************
% ******************************** Metrics
% ****************************************
\subsection{Trustworthy Metrics}

It is not sufficient for systems to just be trustworthy, they must be clearly identifiable as such. By developing useful and informative metrics specific to each trustworthy property, developers, verification engineers, regulators and end users can be better informed of the system's trustworthiness.

At the system level, Floridi suggests the need for agreed-upon metrics for trustworthiness of AI systems and suggests an  AI Trust comparison index, metrics that are needed for benchmarking AI suitability to the public or specific applications~\cite{Floridi2018}. 
Rudas and Haidegger also supports the idea of agreed-upon metrics from the verification community that can be used to ensure reliability of complex autonomous systems~\cite{Rudas2020}. Wang et al. go further and propose a theoretical framework of \emph{tripartite trustworthiness} covering; \emph{to-be trust} (trustfulness of an entity or structure), \emph{to-do trust} (trust in an action or behaviour) and \emph{system trust} (a statistical runtime evaluation of performance) and set out 18 formal definitions~\cite{Wang2020}. 
Garbuk presents the idea of \emph{applied intellimetry} to assess the quality of AI systems by formulating a list of quality characteristics in a functional characteristic vector~\cite{garbuk2018intellimetry}. 
Kaur et al. suggest explainability metrics based on the Euclidean distance between the system output compared to a panel of experts~\cite{kaur2021trustworthy}. 
%
%trustworthiness of computer systems using metrics designed to assess security, trust, resilience and agility~\cite{cho2019stram}

In addition to system level trust, there must be metrics that provide transparency for specific trustworthy qualities. 
Bolster and Marshall proposes the idea of \emph{multi-vector trust metrics} for networks of autonomous systems, indicating that the use of \emph{grey relational analysis}, a theory to describe and model uncertainty, could be beneficial for combining temporally sparse, low fidelity metrics with unknown statistical distributions~\cite{Bolster2014}. 

For data-centric and highly objective measures of trust, operators such as accuracy, precision and recall can be useful for functionality metrics. Questions over what is the ground truth, a sample of the real world, artificially augmented, and if the data is ethically sourced are all additional factors to be considered. We may even be more abstract and use \emph{task completion} - this will confluence with other assessment areas. 
Other, more subjective qualities will use metrics appropriate for the discipline. Qualities such as fairness, cooperation and beneficence may require questionnaires, polls and user feedback to get the correct level of detail. 
%
%\textcolor{red}{Pete and Arianna - can you add to this last point?}

% ****************************************
% ****************** Conflicting Qualities
% ****************************************
\subsection{Conflicting Qualities} \label{sec:conflict}

A trustworthy system could be defined as having sufficient evidence that provides assurance against all trustworthy qualities that are deemed necessary and appropriate for the intended application. But what if maximising one system quality meant reducing another? How should this trade-off be managed and does this lead to competing demands~\cite{Abeywickrama2022}? 
For example, a self-driving vehicle has a performance quality to arrive efficiently at a destination by minimising journey time and is hence optimal at some legal, non-zero speed. However, many safety properties are maximised when the vehicle is at or close to zero speed; a vehicle does no harm if stationary~\cite{harper2021safety}. 
It is possible to imagine a scenario in a busy market square filled with pedestrians and the autonomous vehicle trying to make progress in these conditions. Typically pedestrians will feel more comfortable to be in the road due to the conditions (the market) and cultural norms (consensus with other pedestrians). Maximal driving safety must ensure the vehicle does not come close to any pedestrians but must be traded-off against the need to make progression in the journey.

This type of conflict is conventionally resolved, in the UK at least, with the driver indicating intention by taking small, low speed steps through the crowd, effectively pushing through the crowd with a pseudo-social force effect~\cite{helbing1995social}. 
Although this serves as a good example it does not help with any generalisable rules that can be applied to trading off qualities other than the fact they may be application- and culturally-specific. 

% ****************************************
% ******************* Assessment Framework
% ****************************************
\section{Assessment Framework Vision} \label{sec:AssFramVis}

In this section, the stages of an assessment process are described and illustrated in Fig.~\ref{fig:tas_ver}. Here, the main considerations for trustworthy AS are considered along with: parties that have vested interest in prevention of failure (\emph{trust stakeholders}), how the application will influence the selection and priority of risk, what assessment methods are available and how trustworthiness can be communicated through the use of appropriate metrics. 
From reviewing the literature, a comprehensive spectrum of trustworthy qualities have been derived, which were explored in Section~\ref{sec:astq}, and these qualities form the start of the assessment.

% \subsection{TASQ Categories}

% ****************************************
% ******************************** Stage 1
% ****************************************
\subsection{Stage 1: Trustworthy Qualities, Application Filtering \& Criticality Analysis}
% Which TASQ to include in your application? Are all qualities relevant? How can quality relevance be objectively assessed? What inputs are required from trust stakeholders? How does automation level affect the choice?    

Not all trustworthy qualities are relevant to all applications, so the first stage of the process is to identify the qualities that are relevant to the specified application and filter out those that are irrelevant or inappropriate. 
A criticality assessment is applied at this stage, using a process such as a risk or hazard analysis for each of the selected qualities. 
% not forgetting to assess the impact that automation scope may have on the final outcome, and what potential harm or impact of failing to uphold the specific quality will have on the outcomes of the system and it's users. 
%
The consensus on which qualities to include in the assessment and how important they are will be decided by the \emph{trust stakeholders}, a collection of individuals, groups and organisations who have a vested interest in ensuring the trustworthiness of the AS. Theses stakeholders may include, for example, regulators, developers, end users and incidental users. 
Important factors that the stakeholders will consider are: the scope of automation, how automated the system is or the ambitions it has to be in the future, and the application criticality (including the potential harm from the system not upholding the trustworthy attribute or quality). 
The result from this stage in the process should be a list of qualities for the given application, each with an assigned criticality rating (e.g. high, medium and low risk) that will be used in the assessment stage. 

The main challenge in this area is the need to identify the failure routes of AI and ML systems in comparison to conventional engineering systems. Risk assessments for AS need to include the specific failure modes of AI~\cite{piorkowski2023quantitative} and account for the \emph{brittle} nature of AI models~\cite{druce2021brittle}, or the reliability of predictions made on unseen data~\cite{wing2021trustworthy}. 

%Note: AI failure mode different from conventional engineering. Unseen data causes issues, don't know when a model is confident in answer - VAE could b useful here.

% ****************************************
% ******************************** Stage 2
% ****************************************
\subsection{Stage 2: Identify Standards and Metrics}

Stage 2 of the process is where each quality is assessed according to the appropriate standards and regulations relevant to that property, or the specifications on expected behaviour if such standards are immature or missing. 
At this point, quality metrics should be considered that clearly visualise each of the trustworthiness qualities under scrutiny. 
Many metrics already exist, e.g. precision and recall for predictive analytics, but some will need to be generated and agreed upon by the community, e.g. cooperation, fairness. 
The area of standards and regulations, as discussed in Section~\ref{sec:standards}, is generally immature for AS but is moving rapidly in terms of new publications and frameworks, such as the ratified version of the EU AI act~\cite{EUAIact2021}. 

The challenges in this area are not just the development of standards for AI, but also doing so at a pace that matches  technological advancement. 
Another challenge is the development of appropriate metrics to improve \emph{verifiability}, which is the ease by which a system can be understood and explained, e.g. to a stakeholder or regulator. 

Knowledge of the internal state of the system is often hidden (i.e. a 'blackbox') due to IP and commercial sensitivity, but whitebox access will be essential for certain aspects of trustworthiness assessment. This may not need to reveal sensitive algorithms, but just enough information through observability points in the software architecture could go a long way towards understanding if automated decisions are made for the right reason~\cite{koopman2018toward}. 
Introduction of verifiability at the design stage can pay off in the later assessment and verification stages. An example of this is designing status-indicating LEDs into electronics to visualise certain operational parameters, such as power or network connectivity, which can be clearly understood without need for deeper analysis or tools. 
This is a key challenge to the community and is being addressed through the developments in explainable AI (XAI) that include model layers or separate models that help interpret, for example, image recognition focus~\cite{petsiuk2021black} or general decision making~\cite{danilevsky2020survey, gunning2019xai}.

% ****************************************
% ******************************** Stage 3
% ****************************************
\subsection{Stage 3: Quality Assessment with Corroborative V\&V}

\begin{figure}[]
	\centering
	\includegraphics[width=0.98\linewidth, trim=6.5cm 2.0cm 10cm 3cm, clip]{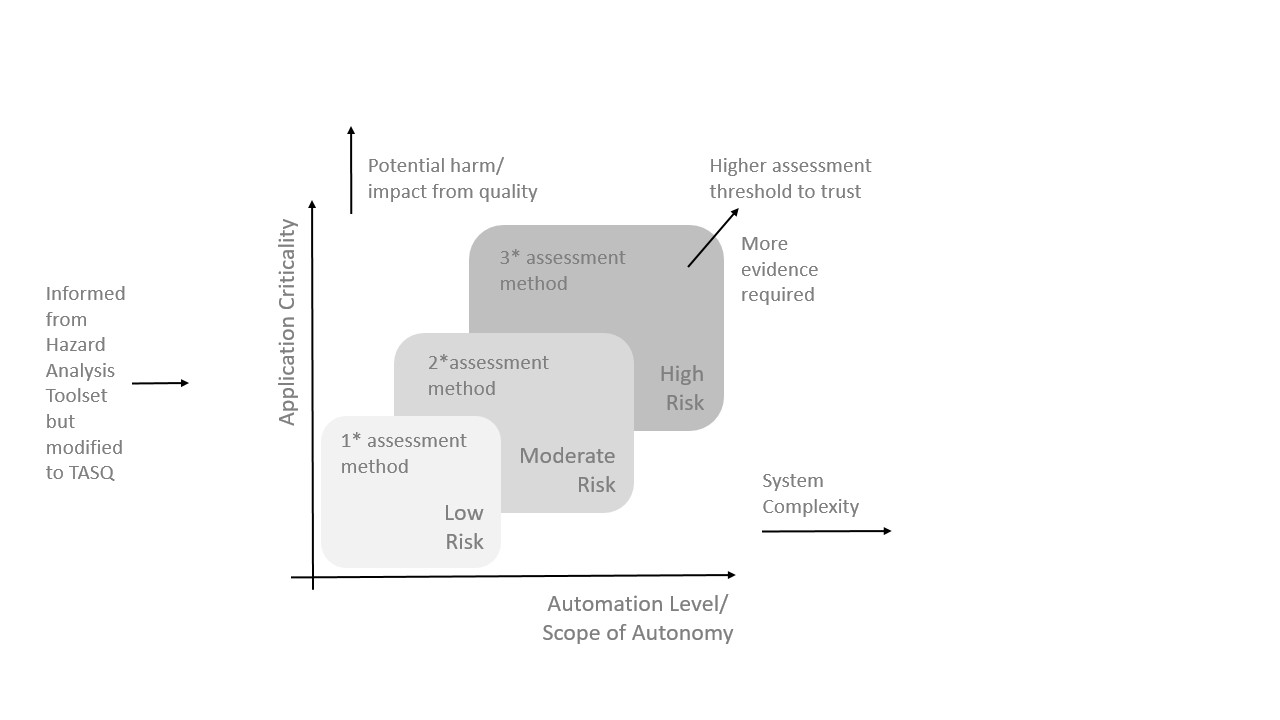}
	\caption{Application criticality and automation level.}
	\label{fig:critauto}
\end{figure}

At Stage 3, each of the qualities from Stage 1 are assessed using one or more verification approaches depending on the risk level identified. Where a higher risk is identified, more evidence should be gathered for that property from independent verification methods, Fig~\ref{fig:critauto}. Mutually consistent, corroborating evidence from independent verification methods should be seen as the gold standard for high-risk trustworthy system properties. 
For example, a quality deemed low-risk may use a single method to provide sufficient evidence of trustworthiness, but for higher-risk categories corroborative evidence from multiple approaches is recommended. 
During each assessment, the trustworthy property will be verified against the respective set of standards, regulations or specifications as appropriate. The quality metrics will be used to visualise the outcomes and give a common language by which the system trustworthiness can be communicated. 

Kress-Gazit et al. state that assessing the trustworthiness of AS can be categorised into four approaches: \emph{synthesis} of correct-by-construction systems, \emph{formal} verification at design time, \emph{runtime} verification or monitoring, and \emph{testing} methods~\cite{kress2021formalizing}. 
An addition to this list may be the inclusion of another verification assessment method called the Principles Based Argument (PBA), that can be used for more subjective qualities such as those in the Ethics category, see ~\cite{Porter2022}. 
%
%Initially each quality will be assessed using either synthesis, testing or formal methods (runtime monitoring is considered in Stage 4). However, the higher the risk criticality then a greater number of methods should be used. 
%
%
%For example, a robot swarm may be monitored for efficiency by monitoring idle individuals and representing this to the tester, regulator or end user as a utilisation score. 

% ****************************************
% ******************************** Stage 4
% ****************************************
\subsection{Stage 4: Assessment Analysis}

Corroborative, mutually-consistent evidence from diverse methodologies provides assurance that is greater in quality than evidence from single sources. 
The assessment analysis should inspect the evidence for each trust quality relative to the risk level based on the application criticality and automation scope. Confidence in the trustworthiness of the system will require more evidence for applications where there is a greater risk level. 
Where appropriate, there may be minimum requirement levels that must be achieved to attain trustworthiness which will be dictated by the appropriate standard for the AI algorithm, application or sector.  %For certain quality categories, for example 98\% accuracy for an image recognition task.
Evidence that fails to corroborate, although does not fail in itself, e.g. sampling different parts of the input space, may be considered equivalent to single-source evidence and therefore, although not a failure, can only provide evidence for lower-risk applications. 
Where there is insufficient availability of independent testing methodology, risks would have to be carried forward and appropriate mitigation plans used during operation. Confidence in trustworthiness can also be gathered from runtime data, although there are obvious dangers with carrying too much risk forward into a live system.

% ****************************************
% ******************************** Stage 5
% ****************************************
\subsection{Stage 5: Visualising Trust}

\begin{figure}[t]
    \centering
    \includegraphics[width=0.98\linewidth, trim=0cm 0cm 0cm 0cm, clip]{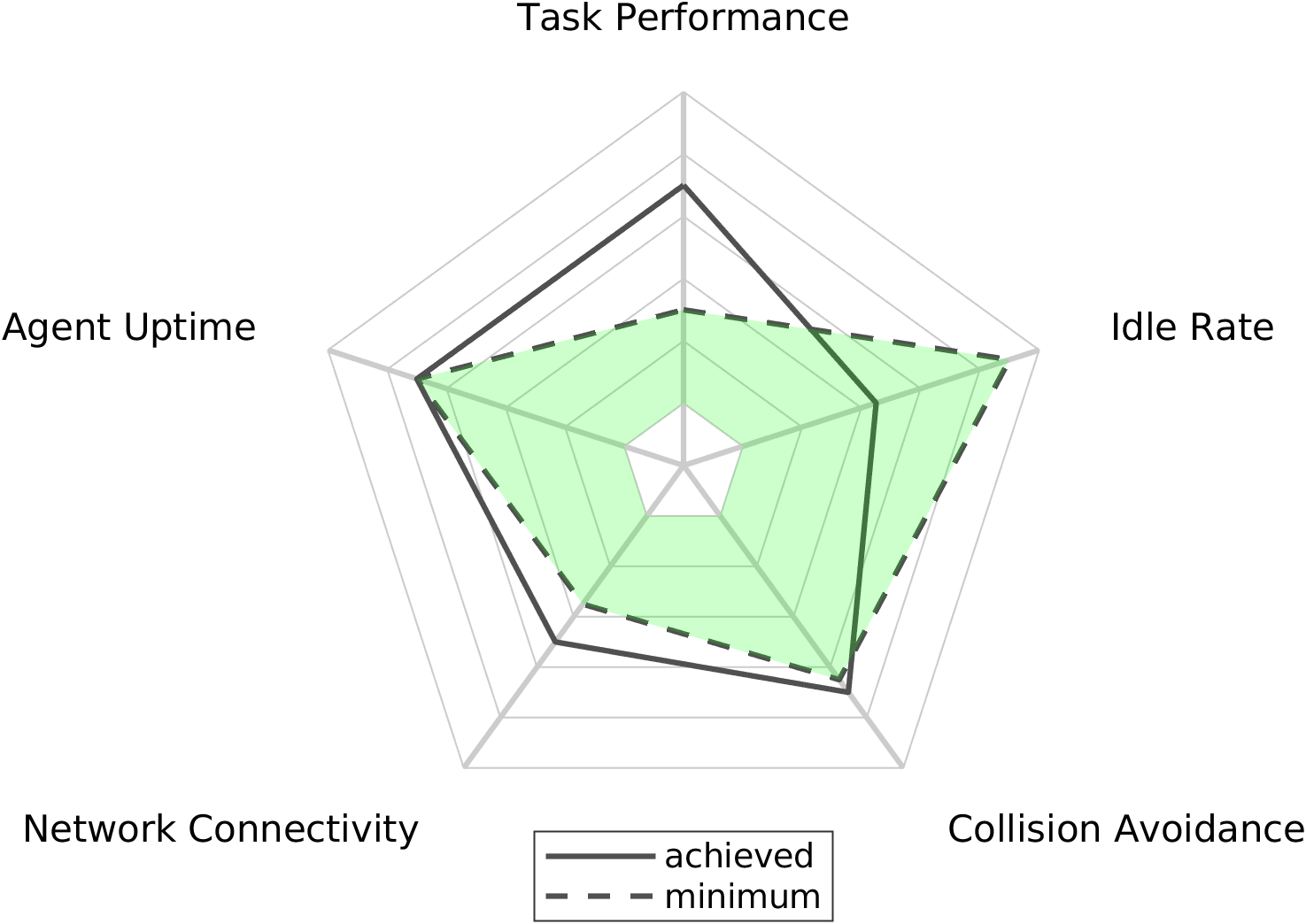}
    \caption{Example of visualising trust in a fictitious swarm robot application.}
    \label{fig:spider}
\end{figure}

Visualising the trustworthiness of the system will be an important aspect of not just presenting objective evidence to developers and regulators, but also to end users who may have a lay understanding of the technical aspects of the system. 
Discussion on what type of information to present to users and how this is done extends beyond the scope of this paper, but is an open challenge to the community. This must be considered as trust visualisation forms the reciprocal part of the trust agreement between user and system. 
However, there are several aspects of visualising trust that should be considered: which categories to display and the format, how the system performed during assessment or currently (if runtime monitoring is used), and a way of indicating the minimum requirements for each quality.

A fictitious example of how trustworthiness of a swarm application could be presented is shown in Fig.\ref{fig:spider}, showing a radar plot of disparate qualities plotted on shared axes. It is clear at a glance that most of the system qualities deemed important, by the trust stakeholders for example, are at or above the minimum threshold required from the associated standards.
Each quality can be seen compared to a minimum level (dashed line) and where the quality exceeds (Task Performance) or fails to reach a minimum level (Idle Rate) alerting the user to areas that may require remediating action (in this case the swarm is under-utilised and scores poorly for idle rate, therefore requiring a reduction in swarm size). 

\subsubsection{Visualising Overall Trust}

A complex, detailed visualisation of trust, as in Fig.\ref{fig:spider}, may be suitable in some cases but at other times it may be more appropriate to present an abstract, higher level visualisation, especially for general public and lay users who are not operators, e.g. purchasing manager. 
As suggested by Floridi~\cite{Floridi2018}, public confidence in AI-based systems could be bolstered with an internationally recognised index for trustworthy AI, such as a \emph{trust comparison index} or AI \emph{star rating}. Much like an internet browser may convince users of their privacy on a web page through some simple means, e.g. a padlock icon, a similar approach could be taken with AS applications. 

A vision of this rating system is presented in Table~\ref{tab:tasq_rating}. 
A star rating is awarded based on the level of corroborative evidence from independent assessment methods that meet the appropriate compliance thresholds when viewed cumulatively across the spectrum of required qualities for the system. 
Essentially, the more diverse the evidence, the greater the trustworthiness rating will be - assuming that assessment outcomes are positive. 

The proposal is that applications with lower risk levels are not required to provide the same level of evidence as higher risk level applications, and therefore the highest burdens of proof are required for higher risk applications.

\subsection{Stage 6: Assessment Outcomes}

The assessment and subsequent analysis should identify qualities that are sufficient and deemed trustworthy and those that fail to meet the requirements. 

\subsubsection{Insufficient Evidence: Update \& Reassessment}

If the analysis shows that the quality fails to meet the minimum specification or there is insufficient evidence that the quality meets the relevant standards, the systems fails on that quality. The system must then pass into an iterative development cycle, where it is updated and then reassessed. 

\subsubsection{Sufficient Evidence: Monitoring}

If there is sufficient evidence that the quality meets the relevant standards, then the system passes that quality. In some applications it may be prudent to enter into a phase of operational assessment where trustworthy metrics are observed in real-time. This \emph{runtime monitoring} allows system trustworthiness to continue to be assessed, which may be required if the system is adaptive to new environments, has evolving functionality, or design stage risk is being carried forward. 
There are challenges associated with runtime monitoring, such as the availability of a suitable oracle~\cite{Leucker2009} to act as a guide for trustworthy operation. This type of operational assurance can be achieved through the use of formal models of the system or AI-subsystem~\cite{dementyeva2022runtime} or by estimating state confidence~\cite{dementyeva2022runtime}.

\begin{table*}[t]
\caption{Trustworthy Autonomous System Quality (TASQ) star rating comparison index}\label{tab:tasq_rating}
\centering
\begin{tabular}{lll}
\toprule
TASQ  &  Assessment & Application \\ 
Rating & Description & Class \\ \midrule

\FiveStar & Single assessment method, meets minimum compliance  & low risk applications only\\
&with standards for low risk applications & \\

\FiveStar\FiveStar & 1-2 assessment methods where appropriate, meets recommended & low-moderate risk applications\\
& compliance level and attempt to calibrate user trust & \\

\FiveStar\FiveStar\FiveStar & evidence from at least 2 diverse assessment methods,  & any risk level applications\\
&meets full compliance guidelines and extensive  &\\
&user trust calibration&\\

\bottomrule
\end{tabular}

\label{tab:tasq_rating}
\end{table*}

\subsection{Future Challenges in Standards} \label{sec:AssFramVis-fut}

Riaz et al.~\cite{Riaz2018} suggest the idea of using social norms and human emotions as a standard by which better self-driving controllers may be developed. This idea sets the way for not just development of higher functioning AS, but also standards of trustworthiness by which they can be judged. Although there is much scholarly work on the theory and modelling of social norms, e.g.~\cite{hechter2001social}, a standard is yet to be published that could be used to objectively assess an autonomous system, and would be subjective to political and cultural influences. 

In some use cases such as driving, legislation on appropriate conduct is presented to society in the form of guidelines such as the UK Highway Code (UKHC) ~\cite{highwayCode}. Often such guidelines must be translated to a computer readable format to act as an appropriate standard or set of assertions~\cite{harper2021safety}, that can be used to assess AS trustworthiness. A similar process will have to be undertaken for other standards which have yet to be defined, e.g. cooperation, fairness or verifiability, to ensure all aspects of trustworthiness can be assessed.

% ****************************************
% ***************************** Assessment
% ****************************************
%
%\subsection{Assessment Methods \& Corroborative Evidence} \label{sec:AssFramVis-AssMthds}
%Gaining reliability assurance of SCASs using testing alone is unfeasible given the often high-dimensional operational state space. Multiple testing methodologies should be employed where appropriate, e.g. verification, falsification and testing, [Harper Corroborative 2022] combining mutually consistent evidence from multiple and diverse assessment methods will raise the confidence in system trustworthiness.

% ****************************************
% ***************************** Conclusion
% ****************************************

\section{Conclusion} \label{sec:conc}

Autonomous Systems are pervasive in our society and the need to demonstrate their trustworthiness has never been more crucial. 
In this paper we have presented an analysis of the literature identifying the key traits or qualities pertaining to trustworthiness of AI and autonomous systems. This literature review has resulted in a \emph{spectrum of trustworthy qualities} which captures the sentiment of the community. These qualities are: Functionality, Reliability, Robustness, Security, Compatibility, Usability, Regulatability and Ethical. Further details of the analysis can be found on our github~\cite{tsl_git}. 
A review of current and emerging standards has been undertaken and although there are only a few that relate directly to AS, there is great interest in the community evidenced by the number of emerging frameworks and draft proposals. 

Some key considerations for trustworthy autonomous systems are considered such as the application criticality and the level of automation scope that the system has. Additionally, trustworthy metrics are discussed alongside their importance in communicating trust with users.

A vision of an assessment process for autonomous and AI-based systems is proposed, and each of the stages of the assessment have been explored relating to current trends in the literature. The stages cover system risk assessment based on application and stakeholder engagement, identification of standards and metrics, assessment and corroborative V\&V, analysis of the assessment and visualisations of trust.

\balance

\section{Acknowledgments}

%The authors would like to thank Kerstin Eder of the University of Bristol for her contribution in discussions and ideas for this paper. 
The work presented has been supported by the UKRI Trustworthy Autonomous Systems Node in Functionality under Grant EP/V026518/1 and Frazer\-Nash Consultancy. 

% ***************************************************
%  Bib
% ***************************************************
%\printbibliography
\bibliography{TASverif.bib} 

\begin{thebibliography}{10}

\bibitem{Murukannaiah2020}
P.~K. Murukannaiah, N.~Ajmeri, C.~M. Jonker, and M.~P. Singh, ``New foundations
  of ethical multiagent systems,'' in {\em Proc. of the 19th International
  Conference on Autonomous Agents and MultiAgent Systems}, AAMAS '20,
  (Richland, SC), p.~1706–1710, International Foundation for Autonomous
  Agents and Multiagent Systems, 2020.

\bibitem{floridi2020gpt}
L.~Floridi and M.~Chiriatti, ``Gpt-3: Its nature, scope, limits, and
  consequences,'' {\em Minds and Machines}, vol.~30, pp.~681--694, 2020.

\bibitem{mediumaiphones}
{\relax Medium}, ``Artificial intelligence in mobile phones.''
  \url{https://medium.com/gobeyond-ai/artificial-intelligence-ai-in-mobile-phones-is-it-a-good-thing-fe044f20ea6c}.
\newblock Accessed: 2022-08-02.

\bibitem{kuo2020towards}
K.~Kuo and D.~Lupton, ``Towards explainability of machine learning models in
  insurance pricing,'' {\em arXiv preprint arXiv:2003.10674}, 2020.

\bibitem{tf_vacuum}
\relax{TensorFlow Blog}, ``Ecovacs robotics: the ai robotic vacuum cleaner
  powered by tensorflow.''
  \url{https://blog.tensorflow.org/2020/01/ecovacs-robotics-ai-robotic-vacuum.html}.
\newblock Accessed: 2022-08-02.

\bibitem{kononenko2001machine}
I.~Kononenko, ``Machine learning for medical diagnosis: history, state of the
  art and perspective,'' {\em Artificial Intelligence in medicine}, vol.~23,
  no.~1, pp.~89--109, 2001.

\bibitem{avci2021review}
O.~Avci, O.~Abdeljaber, S.~Kiranyaz, M.~Hussein, M.~Gabbouj, and D.~J. Inman,
  ``A review of vibration-based damage detection in civil structures: From
  traditional methods to machine learning and deep learning applications,''
  {\em Mechanical systems and signal processing}, vol.~147, p.~107077, 2021.

\bibitem{alpha_fold}
E.~Callaway, ``The entire protein universe: Ai predicts shape of nearly every
  known protein,'' {\em Nature News}, vol.~608, pp.~15--16, 2022.
\newblock Accessed: 2022-08-02.

\bibitem{Abeywickrama2022}
D.~B. Abeywickrama, A.~Bennaceur, G.~Chance, Y.~Demiris, A.~Kordoni, M.~Levine,
  L.~Moffat, L.~Moreau, M.~R. Mousavi, B.~Nuseibeh, S.~Ramamoorthy, J.~O.
  Ringert, J.~Wilson, S.~Windsor, and K.~Eder, ``{On specifying for
  trustworthiness},'' 2022.

\bibitem{althoff2014online}
M.~Althoff and J.~M. Dolan, ``Online verification of automated road vehicles
  using reachability analysis,'' {\em IEEE Transactions on Robotics}, vol.~30,
  no.~4, pp.~903--918, 2014.

\bibitem{CyRes20}
C.~Maple, P.~Davies, K.~Eder, C.~Hankin, G.~Chance, and G.~Epiphaniou, ``Cyres
  -- avoiding catastrophic failure in connected and autonomous vehicles
  (extended abstract),'' 2020.

\bibitem{Lee2004}
J.~D. Lee and K.~A. See, ``{Trust in automation: Designing for appropriate
  reliance},'' {\em Human Factors}, vol.~46, no.~1, pp.~50--80, 2004.

\bibitem{Kohn2021}
S.~C. Kohn, E.~J. de~Visser, E.~Wiese, Y.~C. Lee, and T.~H. Shaw,
  ``{Measurement of Trust in Automation: A Narrative Review and Reference
  Guide},'' {\em Frontiers in Psychology}, vol.~12, no.~October, 2021.

\bibitem{kok2020trust}
B.~C. Kok and H.~Soh, ``Trust in robots: Challenges and opportunities,'' {\em
  Current Robotics Reports}, vol.~1, no.~4, pp.~297--309, 2020.

\bibitem{Chiou2021}
E.~K. Chiou and J.~D. Lee, ``{Trusting Automation: Designing for Responsivity
  and Resilience},'' {\em Human Factors}, 2021.

\bibitem{Floridi2018}
L.~Floridi, J.~Cowls, M.~Beltrametti, R.~Chatila, P.~Chazerand, V.~Dignum,
  C.~Luetge, R.~Madelin, U.~Pagallo, F.~Rossi, B.~Schafer, P.~Valcke, and
  E.~Vayena, ``{AI4People—An Ethical Framework for a Good AI Society:
  Opportunities, Risks, Principles, and Recommendations},'' {\em Minds and
  Machines}, vol.~28, no.~4, pp.~689--707, 2018.

\bibitem{kress2021formalizing}
H.~Kress-Gazit, K.~Eder, G.~Hoffman, H.~Admoni, B.~Argall, R.~Ehlers,
  C.~Heckman, N.~Jansen, R.~Knepper, J.~K{\v{r}}et{\'\i}nsk{\`y}, {\em et~al.},
  ``Formalizing and guaranteeing human-robot interaction,'' {\em Communications
  of the ACM}, vol.~64, no.~9, pp.~78--84, 2021.

\bibitem{Fewster1999}
M.~Fewster and D.~Graham, {\em Software Test Automation Effective use of test
  execution tools}.
\newblock 1999.

\bibitem{trackmenot2009}
H.~Nissenbaum and H.~Daniel, ``Trackmenot: Resisting surveillance in web
  search,'' 2009.

\bibitem{Porter2022}
Z.~Porter, I.~Habli, and J.~McDermid, ``{A Principle-based Ethical Assurance
  Argument for AI and Autonomous Systems},'' pp.~1--39, 2022.

\bibitem{webster2020corroborative}
M.~Webster, D.~Western, D.~Araiza-Illan, C.~Dixon, K.~Eder, M.~Fisher, and
  A.~G. Pipe, ``A corroborative approach to verification and validation of
  human--robot teams,'' {\em The International Journal of Robotics Research},
  vol.~39, no.~1, pp.~73--99, 2020.

\bibitem{schwamm2022}
M.~Schwammberger, C.~Harper, G.~V. Alves, G.~Chance, T.~Pipe, and K.~Eder,
  ``Integrating formal verification and simulation-based assertion checking in
  a corroborative v\&v process,'' {\em arXiv preprint arXiv:2208.05273}, 2022.

\bibitem{Leucker2009}
M.~Leucker and C.~Schallhart, ``{A brief account of runtime verification},''
  {\em Journal of Logic and Algebraic Programming}, vol.~78, no.~5,
  pp.~293--303, 2009.

\bibitem{eder2021cyres}
K.~Eder, ``Cyres: towards operational cyber resilience,'' in {\em Proceedings
  of the 1st International Workshop on Verification of Autonomous \& Robotic
  Systems}, pp.~1--3, 2021.

\bibitem{highwayCode}
\relax{UK Driving Standards Agency}, {\em The Official Highway Code}.
\newblock Her Majestys Stationery Office, 2012.

\bibitem{Bonnefon2016}
J.-F. Bonnefon, A.~Shariff, and I.~Rahwan, ``The social dilemma of autonomous
  vehicles,'' {\em Science}, vol.~352, no.~6293, pp.~1573--1576, 2016.

\bibitem{kraus2022trustworthy}
J.~Kraus, F.~Babel, P.~Hock, K.~Hauber, and M.~Baumann, ``The trustworthy and
  acceptable hri checklist (ta-hri): questions and design recommendations to
  support a trust-worthy and acceptable design of human-robot interaction,''
  {\em Gruppe. Interaktion. Organisation. Zeitschrift f{\"u}r Angewandte
  Organisationspsychologie (GIO)}, pp.~1--21, 2022.

\bibitem{winfield2021ieee}
A.~F. Winfield, S.~Booth, L.~A. Dennis, T.~Egawa, H.~Hastie, N.~Jacobs, R.~I.
  Muttram, J.~I. Olszewska, F.~Rajabiyazdi, A.~Theodorou, {\em et~al.}, ``Ieee
  p7001: A proposed standard on transparency,'' {\em Frontiers in Robotics and
  AI}, p.~225, 2021.

\bibitem{kusters2010}
R.~Küsters, T.~Truderung, and A.~Vogt, ``Accountability: Definition and
  relationship to verifiability,'' {\em Proceedings of the ACM Conference on
  Computer and Communications Security}, pp.~526--535, 2010.

\bibitem{koopman2018toward}
P.~Koopman and M.~Wagner, ``Toward a framework for highly automated vehicle
  safety validation,'' {\em SAE Technical Paper, Tech. Rep}, 2018.

\bibitem{Yazdanpanah2021}
V.~Yazdanpanah, E.~H. Gerding, S.~Stein, M.~Dastani, C.~M. Jonker, and T.~J.
  Norman, ``{Responsibility research for trustworthy autonomous systems},''
  {\em Proceedings of the International Joint Conference on Autonomous Agents
  and Multiagent Systems, AAMAS}, vol.~1, pp.~57--62, 2021.

\bibitem{Hancock2021}
P.~A. Hancock, T.~T. Kessler, A.~D. Kaplan, J.~C. Brill, and J.~L. Szalma,
  ``{Evolving Trust in Robots: Specification Through Sequential and Comparative
  Meta-Analyses},'' {\em Human Factors}, vol.~63, no.~7, pp.~1196--1229, 2021.

\bibitem{Floridi2019}
L.~Floridi, ``{Establishing the rules for building trustworthy AI},'' {\em
  Nature Machine Intelligence}, vol.~1, no.~6, pp.~261--262, 2019.

\bibitem{Thiebes2021}
S.~Thiebes, S.~Lins, and A.~Sunyaev, ``{Trustworthy artificial intelligence},''
  {\em Electronic Markets}, vol.~31, no.~2, pp.~447--464, 2021.

\bibitem{Wing2021}
J.~M. Wing, ``{Trustworthy AI},'' {\em Communications of the ACM}, vol.~64,
  no.~10, pp.~64--71, 2021.

\bibitem{atkinson2012trust}
D.~Atkinson, P.~Hancock, R.~R. Hoffman, J.~D. Lee, E.~Rovira, C.~Stokes, and
  A.~R. Wagner, ``Trust in computers and robots: The uses and boundaries of the
  analogy to interpersonal trust,'' in {\em Proceedings of the Human Factors
  and Ergonomics Society Annual Meeting}, vol.~56, pp.~303--307, 2012.

\bibitem{devitt2018trustworthiness}
S.~Devitt, ``Trustworthiness of autonomous systems,'' {\em Foundations of
  trusted autonomy (Studies in Systems, Decision and Control, Volume 117)},
  pp.~161--184, 2018.

\bibitem{ts_foundation}
\relax{Trustworthy Software Foundation}, ``Ts framework.''
  \url{http://www.tsfdn.org/ts-framework/}.
\newblock Accessed: 2022-08-17.

\bibitem{avizienis2004basic}
A.~Avizienis, J.-C. Laprie, B.~Randell, and C.~Landwehr, ``Basic concepts and
  taxonomy of dependable and secure computing,'' {\em IEEE transactions on
  dependable and secure computing}, vol.~1, no.~1, pp.~11--33, 2004.

\bibitem{jobin2019global}
A.~Jobin, M.~Ienca, and E.~Vayena, ``The global landscape of ai ethics
  guidelines,'' {\em Nature Machine Intelligence}, vol.~1, no.~9, pp.~389--399,
  2019.

\bibitem{nist4}
J.~P. Phillips, C.~A. Hahn, P.~C. Fontana, A.~N. Yates, K.~Greene, D.~A.
  Broniatowski, and M.~A. Przybocki, ``Four principles of explainable
  artificial intelligence,'' Interagency or Internal Reportt NISTIR 8312,
  National Institute of Standards and Technology, sep 2021.

\bibitem{keymolen}
E.~Keymolen, ``Trustworthy tech companies: talking the talk or walking the
  walk?,'' {\em AI Ethics}, 2023.

\bibitem{Fisher2021}
M.~Fisher, V.~Mascardi, K.~Y. Rozier, B.-H. Schlingloff, M.~Winikoff, and
  N.~Yorke-Smith, ``Towards a framework for certification of reliable
  autonomous systems,'' {\em Autonomous Agents and Multi-Agent Systems},
  vol.~35, no.~8, p.~65, 2021.

\bibitem{Jia2021}
Y.~Jia, J.~McDermid, T.~Lawton, and I.~Habli, ``The role of explainability in
  assuring safety of machine learning in healthcare,'' 2021.

\bibitem{Rushby2008}
J.~Rushby, ``Runtime certification,'' in {\em Runtime verification}
  (M.~Leucker, ed.), (Berlin, Heidelberg), pp.~21--35, Springer Berlin
  Heidelberg, 2008.

\bibitem{BS8611}
\relax{British Standards Institute}, ``Bs8611:2016 robots and robotic devices,
  guide to the ethical design and application of robots and robotic systems.''
  Online, 2016.

\bibitem{IEEE-P7001}
A.~Winfield, S.~Booth, L.~A. Dennis, T.~Egawa, H.~Hastie, N.~Jacobs, R.~I.
  Muttram, J.~I. Olszewska, F.~Rajabiyazdi, A.~Theodorou, M.~A. Underwood,
  R.~H. Wortham, and E.~Watson, ``Ieee p7001: A proposed standard on
  transparency,'' {\em Frontiers in robotics and AI}, vol.~8, p.~225, 2021.

\bibitem{IEEE-P7007}
\relax{IEEE Standards Association}, ``7007-2021 – ieee ontological standard
  for ethically driven robotics and automation systems.'' Online, 2021.

\bibitem{FinalP7001}
\relax{IEEE Standards Association}, ``7001-2021 standard for transparency of
  autonomous systems.'' Online, 2022.

\bibitem{IEEE-P7010}
\relax{IEEE Standards Association}, ``7010-2020 – ieee recommended practice
  for assessing the impact of autonomous and intelligent systems on human
  well-being.'' Online, 2020.

\bibitem{24028}
\relax{International Organisation for Standardisation}, ``Iso/iec tr 24028:2020
  information technology — artificial intelligence — overview of
  trustworthiness in artificial intelligence,'' 2021.

\bibitem{24029}
\relax{International Organisation for Standardisation}, ``Iso/iec tr
  24029-1:2021 artificial intelligence (ai) — assessment of the robustness of
  neural networks — part 1: Overview,'' 2021.

\bibitem{cdei}
\relax{Centre for Data Ethics and Innovation}, ``Algorithmic transparency
  recording standard.'' Online, 2021.

\bibitem{Kaakai2022}
F.~Kaakai, K.~Dmitriev, S.~Adibhatla, E.~Baskaya, E.~Bezzecchi, R.~Bharadwaj,
  B.~Brown, G.~Gentile, C.~Gingins, S.~Grihon, and C.~Travers, ``Toward a
  machine learning development lifecycle for product certification and approval
  in aviation,'' {\em SAE Int. J. Aerosp.}, vol.~15, may 2022.

\bibitem{airmf}
\relax{National Institute for Standards and Technology}, ``Artificial
  intelligence risk management framework (ai rmf 1.0).'' Online, 2023.

\bibitem{Hawkins2021}
R.~Hawkins, C.~Paterson, C.~Picardi, Y.~Jia, R.~Calinescu, and I.~Habli,
  ``Guidance on the assurance of machine learning in autonomous systems
  (amlas),'' Guidance Version 1.1, University of York, mar 2021.

\bibitem{Ashmore2021}
R.~Ashmore, R.~Calinescu, and C.~Paterson, ``Assuring the machine learning
  lifecycle: Desiderata, methods, and challenges,'' {\em ACM Comput. Surv.},
  vol.~54, may 2021.

\bibitem{EASA2021}
\relax{European Aviation Safety Agency}, ``Easa concept paper first usable
  guidance for level 1 machine learning applications – proposed issue 01.''
  Online, apr 2021.

\bibitem{Mamalet2021}
F.~Mamalet, E.~Jenn, G.~Flandin, H.~Delseny, and C.~Gabreau, ``White paper
  machine learning in certified systems,'' Research Report hal-03176080, IRT
  Saint Exupery ANITI, 2021.

\bibitem{AFE2020}
\relax{AFE 87 Project Members}, ``Afe 87: Machine learning,'' Final Report
  87-REP-01, Aerospace Vehicle Systems Institute, jun 2020.

\bibitem{UL4600}
D.~Prince and P.~Koopman, ``Ul 4600 technical overview.'' Online, oct 2019.

\bibitem{LNE2021}
\relax{LaboratoireNational de Metrologie et d’Essais}, ``Certification
  standard of processes for ai. design, development, evaluation and maintenance
  in operational conditions,'' Certification Standard~2, LaboratoireNational de
  Metrologie et d’Essais, jul 2021.

\bibitem{fenn2023architecting}
J.~Fenn, M.~Nicholson, G.~Pai, and M.~Wilkinson, ``Architecting safer
  autonomous aviation systems,'' {\em arXiv preprint arXiv:2301.08138}, 2023.

\bibitem{Hawkins22}
R.~Hawkins, M.~Osborne, M.~Parsons, M.~Nicholson, J.~McDermid, and I.~Habli,
  ``Guidance on the safety assurance of autonomous systems in complex
  environments (sace),'' 2022.

\bibitem{ISO29119}
\relax{International Organization for Standardization}, ``{ISO/IEC/IEEE 29119
  Software and systems engineering — Software testing}.'' Online, 2013.

\bibitem{tsl_git}
\relax{Trustworthy System Lab}, ``Tas-verif.''
  \url{https://github.com/TSL-UOB/TAS-Verif}.
\newblock Accessed: 2022-08-22.

\bibitem{Englisch2019}
DIN, ``Din spec 92001-1 artificial intelligence - life cycle processes and
  quality requirements,'' pp.~1--23, 2019.

\bibitem{macrae2021learning}
C.~Macrae, ``Learning from the failure of autonomous and intelligent systems:
  Accidents, safety, and sociotechnical sources of risk,'' {\em Risk analysis},
  2021.

\bibitem{SAEJ3016}
\relax{SAE International}, ``{SAE J3016\_201806 – taxonomy and definitions
  for terms related to driving automation systems for on-road motor
  vehicles.}.'' Online, 2018.

\bibitem{Rudas2020}
I.~Rudas and T.~Haidegger, ``Verification, trustworthiness and accountability
  of human-driven autonomous systems,'' in {\em 2021 IEEE International
  Conference on Autonomous Systems (ICAS)}, pp.~1--1, IEEE, 2021.

\bibitem{Wang2020}
Y.~Wang, S.~Yanushkevich, M.~Hou, K.~Plataniotis, M.~Coates, M.~Gavrilova,
  Y.~Hu, F.~Karray, H.~Leung, A.~Mohammadi, S.~Kwong, E.~Tunstel, L.~Trajkovic,
  I.~J. Rudas, and J.~Kacprzyk, ``{A Tripartite Theory of Trustworthiness for
  Autonomous Systems},'' {\em Conference Proceedings - IEEE International
  Conference on Systems, Man and Cybernetics}, vol.~2020-October,
  pp.~3375--3380, 2020.

\bibitem{garbuk2018intellimetry}
S.~V. Garbuk, ``Intellimetry as a way to ensure ai trustworthiness,'' in {\em
  2018 International Conference on Artificial Intelligence Applications and
  Innovations (IC-AIAI)}, pp.~27--30, IEEE, 2018.

\bibitem{kaur2021trustworthy}
D.~Kaur, S.~Uslu, A.~Durresi, S.~Badve, and M.~Dundar, ``Trustworthy
  explainability acceptance: A new metric to measure the trustworthiness of
  interpretable ai medical diagnostic systems,'' in {\em Conference on Complex,
  Intelligent, and Software Intensive Systems}, pp.~35--46, Springer, 2021.

\bibitem{Bolster2014}
A.~B. Bolster and A.~Marshall, ``{A multi-vector trust framework for autonomous
  systems},'' {\em AAAI Spring Symposium - Technical Report}, vol.~SS-14-04,
  no.~April 2014, pp.~17--19, 2014.

\bibitem{harper2021safety}
C.~Harper, G.~Chance, A.~Ghobrial, S.~Alam, T.~Pipe, and K.~Eder, ``Safety
  validation of autonomous vehicles using assertion-based oracles,'' {\em arXiv
  preprint arXiv:2111.04611}, 2021.

\bibitem{helbing1995social}
D.~Helbing and P.~Molnar, ``Social force model for pedestrian dynamics,'' {\em
  Physical review E}, vol.~51, no.~5, p.~4282, 1995.

\bibitem{piorkowski2023quantitative}
D.~Piorkowski, M.~Hind, and J.~Richards, ``Quantitative ai risk assessments:
  Opportunities and challenges,'' 2023.

\bibitem{druce2021brittle}
J.~Druce, J.~Niehaus, V.~Moody, D.~Jensen, and M.~L. Littman, ``Brittle ai,
  causal confusion, and bad mental models: challenges and successes in the xai
  program,'' {\em arXiv preprint arXiv:2106.05506}, 2021.

\bibitem{wing2021trustworthy}
J.~M. Wing, ``Trustworthy ai,'' {\em Communications of the ACM}, vol.~64,
  no.~10, pp.~64--71, 2021.

\bibitem{EUAIact2021}
\relax{European Commission}, ``Proposal for a regulation of the european
  parliament and of the council laying down harmonised rules on artificial
  intelligence (artificial intelligence act) and amending certain union
  legislative acts.'' Online, apr 2021.

\bibitem{petsiuk2021black}
V.~Petsiuk, R.~Jain, V.~Manjunatha, V.~I. Morariu, A.~Mehra, V.~Ordonez, and
  K.~Saenko, ``Black-box explanation of object detectors via saliency maps,''
  in {\em Proceedings of the IEEE/CVF Conference on Computer Vision and Pattern
  Recognition}, pp.~11443--11452, 2021.

\bibitem{danilevsky2020survey}
M.~Danilevsky, K.~Qian, R.~Aharonov, Y.~Katsis, B.~Kawas, and P.~Sen, ``A
  survey of the state of explainable ai for natural language processing,'' {\em
  arXiv preprint arXiv:2010.00711}, 2020.

\bibitem{gunning2019xai}
D.~Gunning, M.~Stefik, J.~Choi, T.~Miller, S.~Stumpf, and G.-Z. Yang,
  ``Xai—explainable artificial intelligence,'' {\em Science robotics},
  vol.~4, no.~37, p.~eaay7120, 2019.

\bibitem{dementyeva2022runtime}
V.~Dementyeva, C.~Hickert, N.~Sarfaraz, S.~Zanlongo, and T.~Sookoor, ``Runtime
  assurance for intelligent cyber-physical systems,'' in {\em 2022 ACM/IEEE
  13th International Conference on Cyber-Physical Systems (ICCPS)},
  pp.~288--289, IEEE, 2022.

\bibitem{Riaz2018}
F.~Riaz, S.~Jabbar, M.~Sajid, M.~Ahmad, K.~Naseer, and N.~Ali, ``{A collision
  avoidance scheme for autonomous vehicles inspired by human social norms},''
  {\em Computers and Electrical Engineering}, vol.~69, pp.~690--704, 2018.

\bibitem{hechter2001social}
M.~Hechter and K.-D. Opp, {\em Social norms}.
\newblock Russell Sage Foundation, 2001.

\end{thebibliography}
\bibliographystyle{ieeetr}

% ***************************************************
%  Appendix
% ***************************************************
% \input{\pathToSourceFiles/appendix.tex}

\end{document}